\documentclass[letterpaper, 10pt, conference]{ieeeconf}
\IEEEoverridecommandlockouts	

\usepackage{multirow}
\usepackage{lipsum}
\usepackage{amsmath}
\usepackage{amssymb}
\usepackage{subcaption}
\usepackage{graphicx}
\graphicspath{{./figures/}}
\usepackage{glossaries}
\usepackage[table,xcdraw]{xcolor}
\usepackage[ruled,vlined]{algorithm2e}
\usepackage[final]{listings}
\usepackage{longtable}
\usepackage{tikz,xcolor,hyperref}
\definecolor{lime}{HTML}{A6CE39}
\DeclareRobustCommand{\orcidicon}{
	\begin{tikzpicture}
	\draw[lime, fill=lime] (0,0) 
	circle [radius=0.16] 
	node[white] {{\fontfamily{qag}\selectfont \tiny ID}};
	\draw[white, fill=white] (-0.0625,0.095) 
	circle [radius=0.007];
	\end{tikzpicture}
	\hspace{-2mm}
}
\foreach \x in {A, ..., Z}{\expandafter\xdef\csname orcid\x\endcsname{\noexpand\href{https://orcid.org/\csname orcidauthor\x\endcsname}
			{\noexpand\orcidicon}}
}

\definecolor{mygray}{rgb}{0.5,0.5,0.4}
\definecolor{mygreen}{rgb}{0.2,0.5,0.2}
\definecolor{myblue}{rgb}{0.2,0.2,0.9}
\definecolor{mykwdclr}{rgb}{0.2,0.6,0.6}

\lstdefinestyle{mybase} {
	language=[Sharp]C,
	breaklines=true,
	showstringspaces=false,
	basicstyle=\small\tt,
	frame=single,
    numbers=left,
    numbersep=8pt,
	numberstyle=\tiny\color{mykwdclr},
	keywordstyle=\color{myblue},
	keywords=[2]{Mathf,Laser_Scanner,MonoBehaviour,Vector3,RaycastHit,Debug,Color,Physics}, 
	keywordstyle=[2]\color{mykwdclr}, 
	commentstyle=\color{mygreen}\ttfamily
}

\lstdefinestyle{mycsh} {
	style=mybase
}

\newacronym{ITS}{ITS}{Intelligent Transportation Systems}
\newacronym{ROS}{ROS}{Robot Operating System}
\newacronym{SUMO}{SUMO}{Simulation of Urban MObility}
\newacronym{URDF}{URDF}{Unified Robot Description Format}
\newacronym{SLAM}{SLAM}{Simultanous Localization and Mapping}
\newacronym{JSON}{JSON}{JavaScript Object Notation}
\newacronym{LDS}{LDS}{Laser Distance Sensor}
\newacronym{LIDAR}{LIDAR}{Light Detection And Ranging}

\title{\LARGE \bf Mobile Delivery Robots: Mixed Reality-Based Simulation Relying on ROS and Unity 3D}

\author{Yuzhou Liu$^{1}$ \emph{Student Member, IEEE}, Georg Novotny$^{1, 2}$\orcidB \emph{Student Member, IEEE}, \\
Nikita Smirnov$^{1, 3}$\orcidC{} Walter Morales-Alvarez$^{1}$\orcidD \emph{Student Member, IEEE}\\
and Cristina Olaverri-Monreal$^{1}$\orcidE{} \emph{Senior Member, IEEE}%
\thanks{Johannes Kepler University Linz; Chair Sustainable Transport Logistics 4.0, Altenberger Straße 69, 4040 Linz, Austria.
	\texttt{\{yuzhou.liu, georg.novotny, nikita.smirnov, walter.morales\_alvarez, cristina.olaverri-monreal\}@jku.at}}%
\thanks{$^2$ UAS Technikum Wien, Hoechstaedtplatz 6, 1200  Vienna, Austria}\thanks{$^3$ Ural Federal University, Department of Communications Technology, Ekaterinburg, Russia }	
}

\newcommand\copyrighttext{%
  \footnotesize \textcopyright 2020 IEEE. Personal use of this material is permitted. Permission from IEEE must be obtained for all other uses, in any current or future   media, including reprinting/republishing this material for advertising or promotional purposes, creating new collective works, for resale or redistribution to servers or lists, or reuse of any copyrighted component of this work in other works.  DOI: \href{https://ieeexplore.ieee.org/document/9304701}{10.1109/IV47402.2020.9304701} }

\newcommand\copyrightnotice{%
\begin{tikzpicture}[remember picture,overlay]
\node[anchor=south,yshift=10pt] at (current page.south) {\fbox{\parbox{\dimexpr\textwidth-\fboxsep-\fboxrule\relax}{\copyrighttext}}};
\end{tikzpicture}%
}

\begin{document}

\maketitle
\copyrightnotice
\thispagestyle{empty}
\pagestyle{empty}
\captionsetup[figure]{name={Fig.},labelsep=period}
\begin{abstract}
In the context of Intelligent Transportation Systems and the delivery of goods, new technology approaches need to be developed in order to cope with certain challenges that last mile delivery entails, such as navigation in an urban environment. Autonomous delivery robots can help overcome these challenges. 
We propose a method for performing mixed reality (MR) simulation with ROS-based robots using Unity, which synchronizes the real and virtual environment, and simultaneously uses the sensor information of the real robots to locate themselves and project them into the virtual environment, so that they can use their virtual doppelganger to perceive the virtual world.
 
Using this method, real and virtual robots can perceive each other and the environment in which the other party is located, thereby enabling the exchange of information between virtual and real objects.

Through this approach a more realistic and reliable simulation can be obtained. 
Results of the demonstrated use-cases verified the feasibility and efficiency as well as the stability of implementing MR using Unity for \gls{ROS}-based robots. 

\end{abstract}

\section{Introduction}
\label{sec:introduction}

Advanced applications that rely on Intelligent Transportation Systems (ITS) combine Information and Communication Technologies (ICT) in a connected environment in which sensors acquire information relevant for services related to different modes of transport,  contributing to their more efficient and sustainable use. Before testing the systems in a real world environment, it is essential that the system performance is validated in a controlled environment using simulation frameworks. \\
In the context of transport logistics, the growth of e-commerce has increased customer demand and new technology approaches need to be developed to cope with the challenges occurring during the final step of the delivery process from a facility to the end user (last mile). These challenges refer, for example, to navigation in an urban environment, including parking and its effect on fuel consumption, fuel costs and C02 emissions. Autonomous delivery robots can help overcome these challenges.\\
We present a method for implementing mixed reality applications for Robot Operating System \cite{ROS} \gls{ROS}-based mobile robots in the open source game engine Unity 3D \cite{helgason2018unity}, thus extending the work presented in \cite{hussein2018ros}. Unity 3D has been widely used in the development of ITS by connecting it to advanced 3D modeling software for  urban environments including CityEngine~\cite{cityengine}, \gls{SUMO}~\cite{SUMO.}, and \gls{ROS} \cite{biurrun2017microscopic, hussein2018coautosim, artal2019, olaverri2018implementation, olaverri2018connection}. \\
The connection of Unity and \gls{ROS} can greatly simplify the development and simulation process of autonomous delivery robots. \\
In this work, we contribute to the field of multirobot simulation by combining a simulated world with the physical world. This approach makes it possible to use a small number of real robots in combination with a large number of virtual robots to perform a multi-robot simulation in a real environment and therefore obtain more realistic and reliable simulation results while improving the efficacy of the entire system.\\
This proposed visualization of robot models and data, also described as the integration of physical objects into a virtual environment (Augmented Virtuality (AV)), can be seen as a special instance of Mixed Reality (MR) \cite{milgram1994taxonomy}. According to \cite{hoenig2015mixed} MR has to be able to spatially map physical and virtual objects alongside one another in real-time in a fusion of the physical world and the virtual world to create new environments and visualizations in which physical and virtual objects coexist and interact in real-time.
Therefore, our work may help lay the foundation for the development of MR applications for mobile delivery robots in ITS.\\
The remainder of this paper is organized as follows: Section \ref{sec:relatedwork} introduces the previous works in the field. The methodology to use Unity to create mixed reality applications for \gls{ROS}-based robots is presented in Section \ref{sec:visualization}. Afterwards, the validation experiments are evaluated in Section \ref{sec:evaluation}. Finally, Section \ref{sec:conclusion} concludes the paper and presents possible future work.

\section{Related Work}
\label{sec:relatedwork}
Before implementing MR applications for \gls{ROS}-based robots in Unity, it is first necessary to establish reliable communication between \gls{ROS} and Unity. Several libraries have been written to integrate Unity with \gls{ROS}. For example, the $rosbridgelib$ uses $rosbridge$ \cite{mace2018ros} for communication and defines messages in Unity to establish bidirectional communication at a later point \cite{thorstensen2018ros}. The approach presented in this paper adheres to this library and added several scripts to extend its support for different \gls{ROS} message types.\\
It is worth mentioning that Unity is used as a full virtual simulation tool rather than a mixed reality platform, examples of the latter being in the modeling and simulation of the humanoid robot in \cite{mattingly2012robot} and the unmanned aerial vehicle (UAV) project introduced in \cite{hu2016rosunitysim}.

As demand changes and mixed reality technologies mature, applications similar to ours in this area are expected to increase.
Related research that laid the foundation for our work is for example the implementation of a \gls{ROS}-based robot controlled by Unity and VR devices~\cite{codd2014ros}. The authors successfully imported \gls{ROS} map information into Unity and created a new scene accordingly, which also enabled the visualization of the robot in Unity to some extent. The paper described the method for sending messages to control the robot through Unity, but the acquisition and visualization of laser scan data in Unity was not mentioned, so that the effect of the process on the final results could not be verified. \\
In order to solve the problem of matching the robot's localization and Unity scene, we refer to the work in \cite{roldan2019multi}. The authors also connected Unity and \gls{ROS} and accomplished the synchronization of real and virtual robot, but Unity was only used as a control medium to transfer commands from the gaming device to \gls{ROS}. 

A similar method was used in \cite{hoenig2015mixed} to create a mixed reality implementation for multiple UAVs based on \gls{ROS} and Unity. The UAVs' positions were tracked through an additional Virtual-Reality Peripheral Network (VRPN) protocol, which is a device-independent and network-transparent system designed specifically for VR peripherals \cite{taylor2001vrpn}. \gls{ROS} was only used to control UAVs, but not connected with Unity.\\
In a further work~\cite{bultmann2017multi} the cooperative \gls{SLAM} and map merging methods were used so that 3 robots could derive their relative position and orientation in the same environment without knowing their initial position. This implementation solved real-time self-localization issues. \\
Another work \cite{nasir2014cooperative} provided a more detailed discussion of similar research, further validating the feasibility of the MR approach and providing a more complete and accurate mathematical framework. Furthermore, in \cite{walter2004experimental} and \cite{bultmann2017multi} the authors adopted cooperative \gls{SLAM} and demonstrated the detection performance of the poses of each robot in the group. In the last work the authors additionally used ROS to share robot information related to position, rotation and scale.\\
In the work presented here we have successfully implemented mixed reality applications using only Unity and ROS and a smart mobile device. At the same time, we successfully enabled the real robot's ability to perceive both the virtual environment and other real-world and virtual robots.

\section{Matching Virtual and Physical Worlds (Mixed Reality)}
\label{sec:visualization}

This section explains in detail the process of matching the virtual and physical worlds using \gls{ROS} and Unity in two stages.
\begin{enumerate}
\item In the first stage, Unity3D is used to visualize and project the robot models from the virtual environment onto the real world. 
\item Afterwards, in a second step, we use \gls{ROS} to share the sensor data and project the current status of the real robots to their virtual doppelgangers in the virtual environment.
\end{enumerate} 

\subsection{Visualization in Unity}
We explain next the concepts required to apply the proposed method. 

\subsubsection{AR Foundation}
We relied on the Unity 3D AR foundation package for implementing augmented reality functions. The package supports devices such as smart phones or tablets. 
We mainly used its planar surface detection function to overlay the virtual scene on the detected plane and make it match the physical surroundings. Its world tracking function made it possible to keep all the information related to the \textit{transform} component that is part of each object within the Unity Scene, and this function is then used to store and manipulate the position, rotation and scale of the object~\cite{helgason2018unity}. 

\subsubsection{Modeling and Coordinate Systems}

Although the AR foundation can provide both horizontal and vertical plane detection and even simple spatial mapping, in order to obtain better performance, it is still required to model the environment manually. \\
Since we used robots that were equipped with 2D Lidars, we relied on the method in \cite{codd2014ros} that used a scanned map of the robots for modeling the environment. But instead of just using point clouds to represent the map we extended the approach to interact with the virtual scene by using models as game objects to rebuild the environment with collider components that rely on Unity’s built-in collision system.\\

By importing the map and using the polygon building tool provided by Unity, we can quickly obtain a rough model of the desired environment, and then manually adjust to get the final, more accurate model. \\
For AR applications, the scaling of the models is very important. In theory, one unit length in Unity is equivalent to one meter in reality. We adjusted our model to a scale that matches reality by measuring and calculating proportions. In Figure \ref{fig:compare} a properly scaled robot model in comparison with the size of a real robot is depicted.\\
Finally, we matched the coordinate systems in Unity and \gls{ROS}. As Unity's game objects adopt a hierarchy structure, by arranging the hierarchies, we were able to obtain a local coordinate system that was not affected by scale change and that coincided with the \gls{ROS} zero point.

\subsubsection{Installing to Device and Testing}

\begin{figure}[ht]
    \centering
    \includegraphics[width=0.49\textwidth]{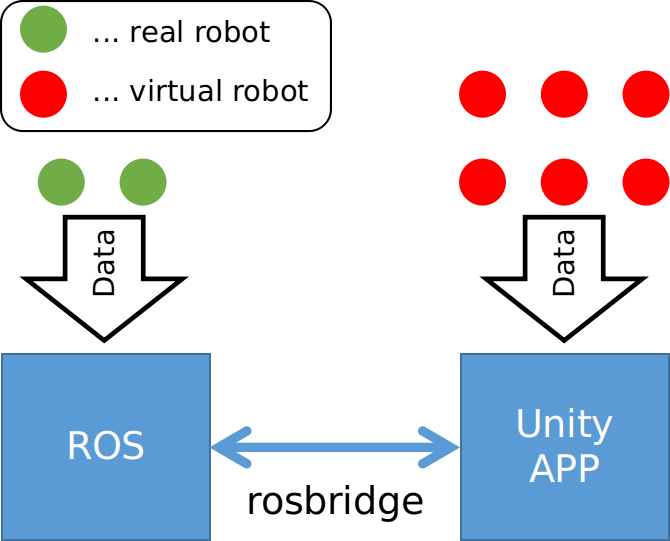}
    \caption{Overview of the communication flow}
    \label{fig:communication}
\end{figure}

\begin{figure}[!b]
	\centering
	\begin{subfigure}[b]{0.49\textwidth}
	    \centering
    	\includegraphics[width=\textwidth]{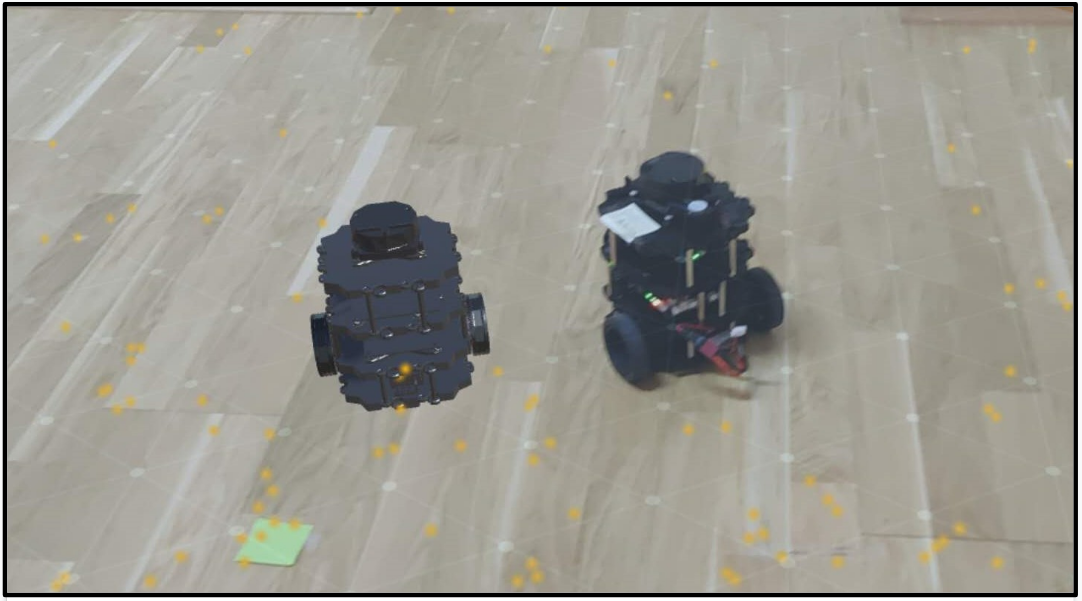}
    	\caption{Properly scaled virtual robot model (left) located on the detected plane and compared with the real robot (right)}
    	\label{fig:compare}
    \end{subfigure}
	\begin{subfigure}[b]{0.49\textwidth}
    	\centering
    	\includegraphics[width=\textwidth]{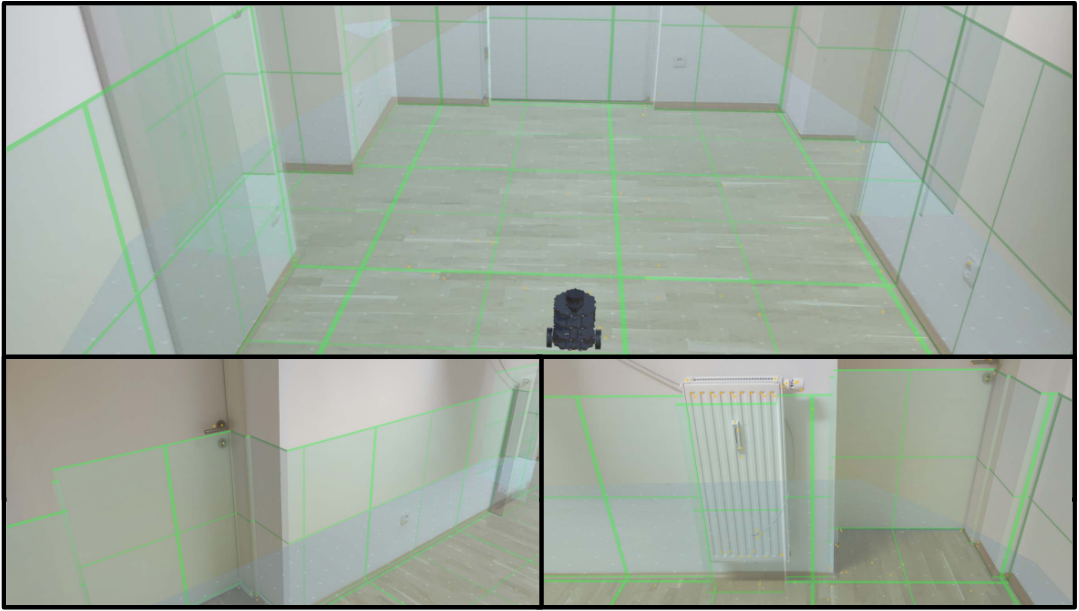}
    	\caption{Illustration of matching the virtual scene with the real world, as well as the proportional relationship between the virtual robot model and the environment}
    	\label{fig:matchevn}
    \end{subfigure}
    \caption{Overview of the virtual robot inside the virtual and real environment}
\end{figure}

Because the sensors on the mobile device are needed to complete the AR function, we implemented an application (APP) and installed it on an Android smart phone. When the APP started running, it was necessary to move the camera of the mobile device to complete the plane detection of the entire environment. At this time, Unity was not connected to ROS. \\
Next, according to the previously written scripts, the experimental scene was initialized by tapping the screen. The manipulations press and drag with fingers  could be used to fine-tune the \textit{transform} of the environment. The connection between Unity and ROS was then established at the same time as the experimental scenario was initialized. The proposed communication flow is depicted in Figure \ref{fig:communication}.
The effect of matching the virtual scene with the real world, as well as the proportional relationship between the virtual robot model and the environment are visualized in Figure \ref{fig:matchevn}.

\subsection{Self-Localization and Data Sharing within ROS}

\begin{figure}[ht]
	\centering
	\begin{subfigure}[b]{0.49\textwidth}
		\centering
		\includegraphics[width=\textwidth]{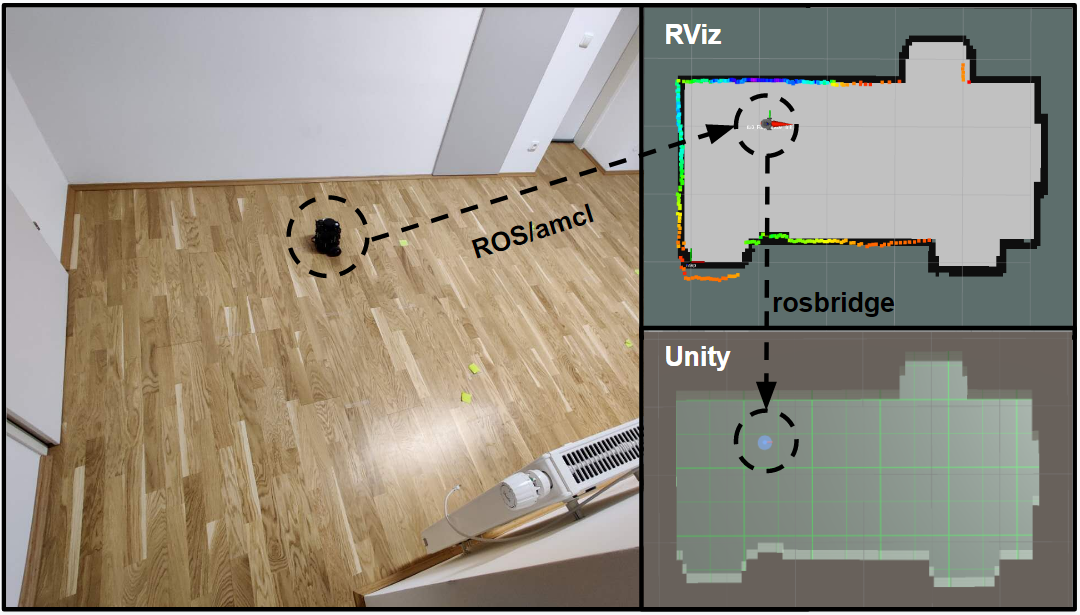}
		\caption{Projection of a real robot into a virtual space}. 
		\label{fig:show}
	\end{subfigure}
	\begin{subfigure}[b]{0.49\textwidth}
		\centering
		\includegraphics[width=\textwidth]{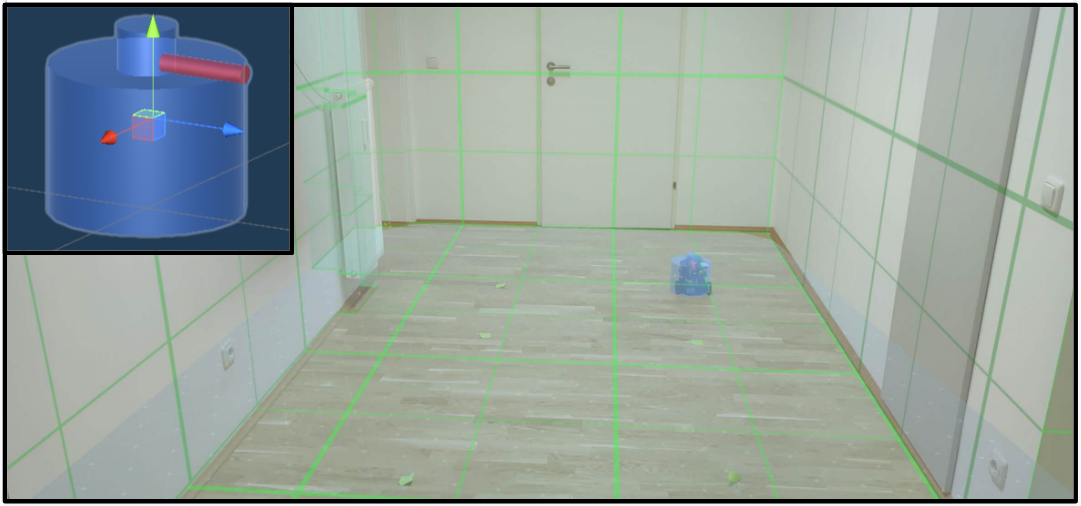}
		\caption{ Augmented reality view resulting from the proposed approach}
		\label{fig:finalview}
	\end{subfigure}
	\caption{Visualization of a real robot in a virtual environment as a result of the matching of scaling and localization}
	\label{fig:projection}
\end{figure}

As previously mentioned we used ROS to share the robot \textit{transform} information. To this end we adhered
to the $amcl$ package, which is used as a probabilistic localization system for robots moving in 2D \cite{amcl2019ros}, so that their location and orientation can be recorded to be later shared within the ROS $tf$ system. 
Since in our approach the virtual and real scenario are completely matched with the map, the virtual and real robots can obtain the relative positions of each other. In fact, after having synchronized the coordinate systems, a single robot can also be accurately mapped into the virtual space.\\
By combining the AR foundation approach with the described method, we can then visualize the position of a real robot in the virtual environment and in the AR scene. Figure \ref{fig:show} illustrates how a real robot is projected into a virtual space, the virtual environment and objects can then be embedded in the real world using mobile devices. After correctly projecting a real robot into a virtual space, its doppelganger can be used to emit another set of laser scans. By sending the scanned data back to ROS the real robot can then perceive virtual objects. The final AR view is depicted in Figure \ref{fig:finalview}. As shown in the upper left corner of the picture, instead of using the virtual robot model, we used an additional translucent model as the doppelganger of the real robot for better visualization purposes. 

\section{System Evaluation and Results}
\label{sec:evaluation}

To evaluate the implementation method, we designed two use-cases as described in this section.

\subsection{Robot Following Behavior}
\begin{figure}[ht]
	\centering
	\includegraphics[width=0.3\textwidth]{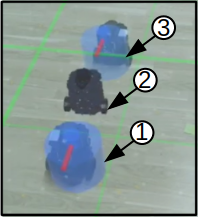}
	\caption{Illustration of the performed robot-following behavior experiment\\ \small{1) leader robot; 2) real follower robot; 3) virtual follower robot}}
	\label{fig:followers}
\end{figure}

We developed an algorithm to test the feasibility of our approach by implementing a scenario in which a leading real robot (leader robot) was followed by a virtual robot (virtual follower robot) that in turn was followed by a real robot (real follower robot) as Figure \ref{fig:followers} illustrates. \\
The leader robot was required to perform self-localization, thereby projecting its position into a virtual environment and generating a doppelganger. 
The virtual follower robot was represented in the Unity environment, so that it could directly obtain its own \textit{transform} information and then transfer it to ROS for sharing.
In order for this second robot to perceive the leader doppelganger in the virtual environment we implemented a virtual laser scan.
The real follower robot in the third position was then required to perform  self-localization by scanning and measuring the environment, to be projected into the virtual space. Then using the laser scan information of the doppelganger in the virtual environment, the real follower robot was able to perceive the  virtual robot in front and then follow its movement.

Through this use case, we successfully verify the feasibility of using Unity to perform MR simulation for ROS-based robots. The transform information of the real robot is accurately projected into the virtual space. Mutual awareness between real and virtual robots has also been successfully established.\\
In this use case, each robot's actions are determined by its own program, rather than an overall control algorithm, which means that the robots' transform information is not fully shared between every robot, and each robot can only sense the target in front of it based on its own sensors. \\
In many cases in the real world, robots are controlled by a central system. In order to verify the performance of this implementation for this central system, we introduce a second use-case in the next subsection.

\subsection{Multi-robot Roundabout Entry and Exit Behavior}

\begin{figure}[ht]
	\centering
	\includegraphics[width=0.49\textwidth]{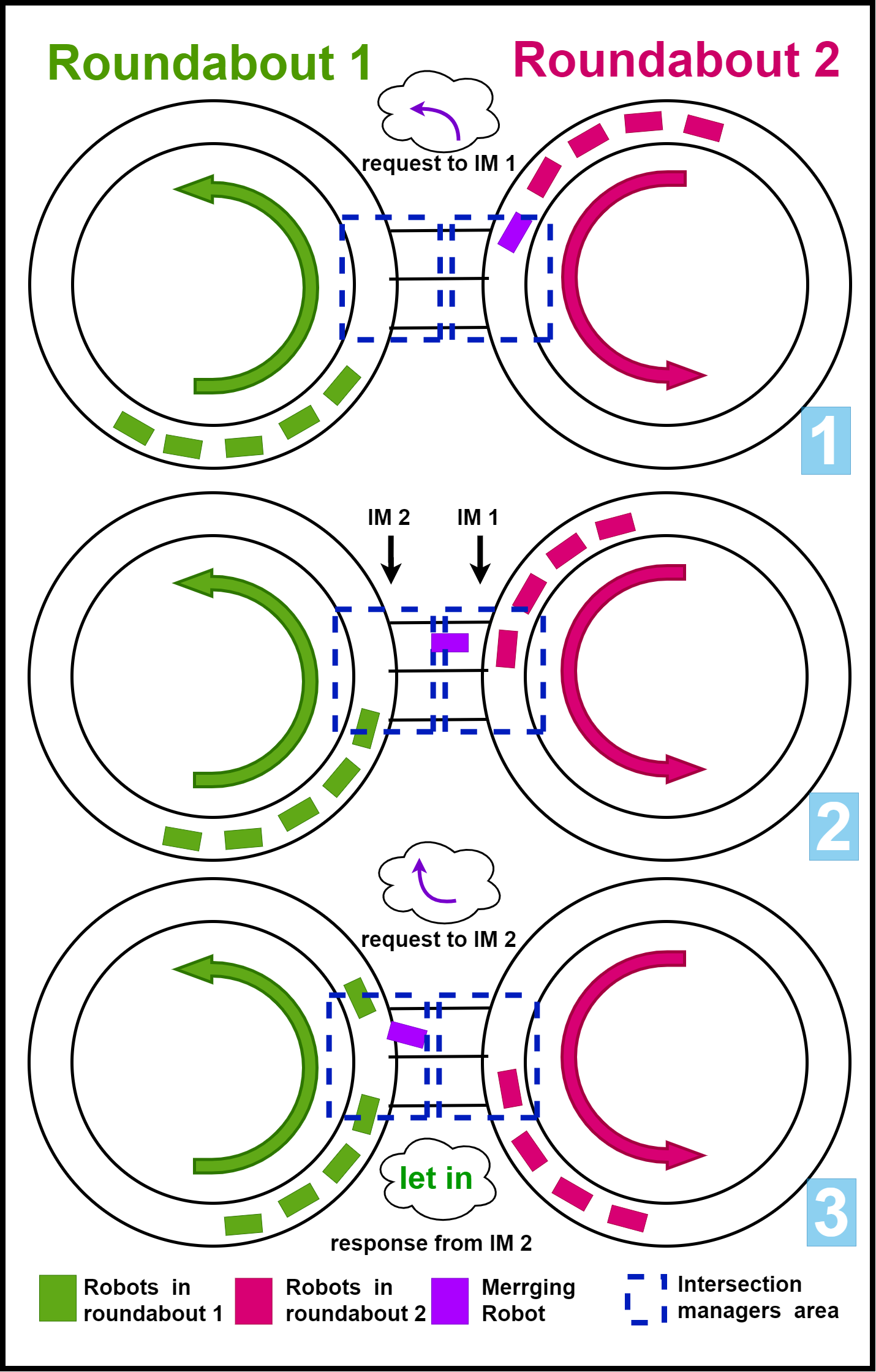}
	\caption{Scenario description; 1: merging robot sends a request to leave from roundabout 2; 2: merging robot enters into an intersection queue; 3: merging robot sends a request to join roundabout 1 and after that robots from roundabout 1 get a message to let in the merging robot}
	\label{fig:roundabouts}
\end{figure}

To test the performance of a multi-robot hybrid simulation we developed a use case with eight Turtlebot3 \cite{Turtlebot}, two of which were real and six virtual. Each robot obtained its pose through ego localization and shared it with the others.
The selected test scenario consisted of two roundabouts with one lane each that were connected by a two-lane road (one lane for each direction), as visualized in Figure \ref{fig:roundabouts}. 

Traffic was generated as follows: 4 robots were placed on each roundabout forming a traffic queue. Each robot was initialized with a unique speed and controlled to follow the lane of the roundabout using a PID controller. Then a manual request was sent to a robot to join the other traffic queue. To do so, they needed to leave the current queue, drive along the two-lane road and make a safe turn onto the new roundabout. Notice that this scenario was designed that only one robot from one queue can leave per cycle.   \\
Since the robots exhibit different speeds, two algorithms were used to control the current velocity. To avoid collisions while being on the roundabout, a basic linear control concept was applied to alter the velocity of a robot  based on the distance acquired from the simulated \gls{LIDAR} sensors inside of Unity. Traffic at the entrance and exit of the roundabout, on the other hand, was managed by an autonomous intersection manager (IM). Notice that each roundabout has it's own IM to handle the requests at their intersection. While robots are inside the IM area they send their current velocity and type of movement (exit intersection, follow traffic queue, enter intersection) to the IM which then handles the traffic as described in algorithm \ref{alg:IM}.

\begin{algorithm}[!h]
	\SetKwInOut{Input}{input}\SetKwInOut{Output}{output}
	\Input{velocity, movement}
	\Output{velocity}
	\BlankLine
	queue $\leftarrow$ new list\\
	\For{ i $\leftarrow$ 0 to \# robots in IM area}{
		queue.append(i)
	}
	\While{True}{
		\For{i $\leftarrow$ to \# queue - 1}{
			pose = get\_pose\_of\_robots(queue[i])\\
			time = calculate\_time\_to\_intersection(pose)\\
			\If{possible\_collision(time, queue[i+1])}{
				adjust\_velocity\_of\_robot(i+1)
			}\Else{return velocity}	
			queue.remove(queue[i])
		}
	}
	\caption{Intersection Manager\label{alg:IM}}
\end{algorithm}

\begin{figure}[ht]
	\centering
	\includegraphics[width=0.49\textwidth]{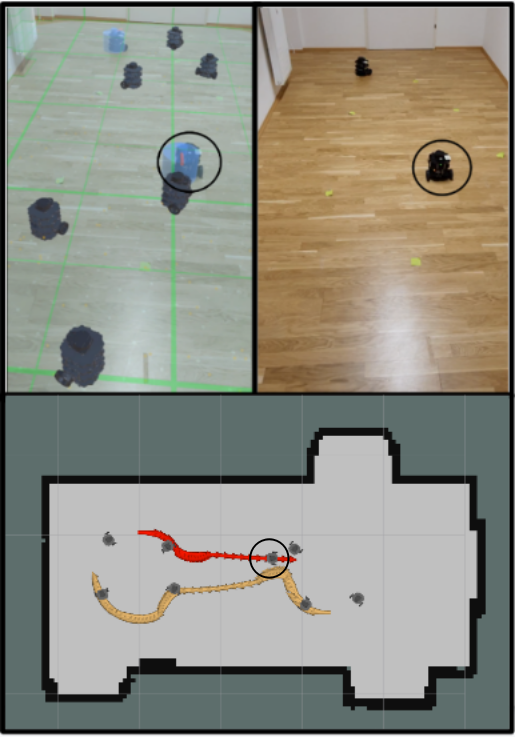}
	\caption{Illustration of the experiment performed, in which a robot waits to be joined in other roundabout by IM; Left: mixed reality view with virtual and real robots (in blue); Right: real world view in which the virtual robots are not visible; Bottom: visualization inside rviz, trajectories of merging robots are colored in red and orange}
	\label{fig:overview}
\end{figure}

After receiving a request, the manager adds the robot's request to a queue and handles the incoming robot requests based on the FIFO principle. The current pose and velocity of each first robot is used to calculate the time to intersection (TTI), utilizing a differential drive kinematic motion model, and calculates possible conflicts for the following robots in the queue based on this. If a potential conflict is detected (i.e. collision), the manager adjusts the speed of the robot whose request arrived after the current first robot. In case of no conflict, the value returned by the manager is the original speed of the robot. Finally the IM removes the current request from the queue and starts over.
Figure \ref{fig:overview} visualizes the test scenario. As it can be seen when comparing the left and right images the IM updated the current velocity of the real robot that was marked within a circle in the figure, so that it would not crash into the simulated robot in the roundabout.

Through a more sophisticated simulation, we verified the stability of the approach. Whether virtual or real, the position and orientation information of each robot can be accurately updated in real-time, and the mutual perception can help them avoiding collisions with each other. Meanwhile, this approach of combining a simulated world with the physical world makes it possible to perform a large-scale simulation that integrates small-scale field tests, thus obtaining more realistic and reliable simulation results than in pure virtual environments and improving the efficacy of the entire system. \\
\section{Conclusion and Future Work}
\label{sec:conclusion}

The work we presented describes a method relying on mixed reality to be applied in the testing of delivery robots behavior and interaction in a realistic simulation framework. Two concrete simulation examples were described, in which the visualization was managed by Unity 3D and the data shared by the robots was managed by the self localization algorithm that was implemented in \gls{ROS}. 

The results from the use cases showed the feasibility and efficiency of the Unity-based MR simulation for ROS-based robots. The matching of virtual and real environments was established, the mutual perception of virtual and real robots was successfully achieved as they were following each other. By using multiple virtual and real robots to simulate complex behaviors, we also successfully verified the stability of the proposed system. At the same time, due to the powerful extend-ability of Unity itself, integrating designed MR use cases into other large-scale simulations is also possible.

Future work will aim at performing spatial mapping dynamically to improve the universality of the MR applications and integrating the results into the ``autonomous vehicles'' option of the 3DCoAutoSim simulation framework~\cite{hussein2018coautosim}. \\

\section*{ACKNOWLEDGMENT}
This work was supported by the Austrian Ministry for Climate Action, Environment, Energy, Mobility, Innovation and Technology (BMK) Endowed Professorship for Sustainable Transport Logistics 4.0.

\bibliographystyle{IEEEtran}
\bibliography{paper}
\end{document}